\documentclass[sigconf]{acmart}
\AtBeginDocument{%
  }

\setcopyright{acmlicensed}
\copyrightyear{2018}
\acmYear{2018}
\acmDOI{XXXXXXX.XXXXXXX}
\acmConference[Conference acronym 'XX]{Make sure to enter the correct
  conference title from your rights confirmation email}{June 03--05,
  2018}{Woodstock, NY}
\acmISBN{978-1-4503-XXXX-X/2018/06}

\renewcommand\footnotetextcopyrightpermission[1]{}
\usepackage{multirow}

\begin{document}

\title{Parser-Oriented Structural Refinement for a Stable Layout Interface in Document Parsing}

\author{Fuyuan Liu}
\authornote{Equal contribution.}
\affiliation{%
  \institution{Unisound AI Technology Co., Ltd.}
  \country{China}
}

\author{Dianyu Yu}
\authornotemark[1]
\affiliation{%
  \institution{Unisound AI Technology Co., Ltd.}
  \country{China}
}

\author{He Ren}
\affiliation{%
  \institution{Unisound AI Technology Co., Ltd.}
  \country{China}
}

\author{Nayu Liu}
\affiliation{%
  \institution{School of Computer Science and Technology, Tianjin University}
  \country{China}
}

\author{Xiaomian Kang}
\affiliation{%
  \institution{MAIS, Institute of Automation, Chinese Academy of Sciences}
  \country{China}
}

\author{Delai Qiu}
\affiliation{%
  \institution{Unisound AI Technology Co., Ltd.}
  \country{China}
}

\author{Fa Zhang}
\affiliation{%
  \institution{Unisound AI Technology Co., Ltd.}
  \country{China}
}

\author{Genpeng Zhen}
\affiliation{%
  \institution{Unisound AI Technology Co., Ltd.}
  \country{China}
}

\author{Shengping Liu}
\affiliation{%
  \institution{Unisound AI Technology Co., Ltd.}
  \country{China}
}

\author{Jiaen Liang}
\affiliation{%
  \institution{Unisound AI Technology Co., Ltd.}
  \country{China}
}

\author{Wei Huang}
\affiliation{%
  \institution{Unisound AI Technology Co., Ltd.}
  \country{China}
}

\author{Yining Wang}
\authornote{Corresponding author.}
\email{wangyining@unisound.com}
\affiliation{%
  \institution{Unisound AI Technology Co., Ltd.}
  \country{China}
}

\author{Junnan Zhu}
\authornotemark[2]
\email{junnan.zhu@nlpr.ia.ac.cn}
\affiliation{%
  \institution{MAIS, Institute of Automation, Chinese Academy of Sciences}
  \country{China}
}

\renewcommand{\shortauthors}{Liu et al.}

\begin{abstract}
Accurate document parsing requires both robust content recognition and a stable parser interface. In explicit Document Layout Analysis (DLA) pipelines, downstream parsers do not consume the full detector output. Instead, they operate on a retained and serialized set of layout instances. However, on dense pages with overlapping regions and ambiguous boundaries, unstable layout hypotheses can make the retained instance set inconsistent with its parser input order, leading to severe downstream parsing errors. To address this issue, we introduce a lightweight structural refinement stage between a DETR-style detector and the parser to stabilize the parser interface. Treating raw detector outputs as a compact hypothesis pool, the proposed module performs set-level reasoning over query features, semantic cues, box geometry, and visual evidence. From a shared refined structural state, it jointly determines instance retention, refines box localization, and predicts parser input order before handoff. We further introduce retention-oriented supervision and a difficulty-aware ordering objective to better align the retained instance set and its order with the final parser input, especially on structurally complex pages. Extensive experiments on public benchmarks show that our method consistently improves page-level layout quality. When integrated into a standard end-to-end parsing pipeline, the stabilized parser interface also substantially reduces sequence mismatch, achieving a Reading Order Edit of 0.024 on OmniDocBench.
\end{abstract}




\maketitle

\section{Introduction}

Document parsing provides the structural bridge between visually rich document pages and downstream language-centric applications such as retrieval, summarization, question answering, and knowledge extraction. Recent benchmarks show that parsing quality depends not only on text recognition accuracy but also on whether the system can faithfully recover page structure, including layout grouping and reading order \cite{omnidochbench}. In practice, even when most text content is recognized correctly, errors in layout retention or ordering can still corrupt the final Markdown or JSON output.

Existing document parsing systems can be broadly divided into two categories. The first category consists of end-to-end parsers that directly generate structured outputs from page images, as exemplified by FireRedOCR \cite{fireredocr}, HunyuanOCR \cite{hunyuanocr}, and DeepSeekOCR \cite{deepseekocr}. The second category follows a pipeline design with an explicit Document Layout Analysis (DLA) stage before downstream parsing. Representative systems in this category include PaddleOCR-VL-1.5 \cite{paddleocrvl15}, GLM-OCR \cite{glmocr}, MinerU2.5 \cite{mineru25}, and Youtu-Parsing \cite{youtuparsing}. Although end-to-end parsers have advanced rapidly, pipeline systems remain attractive for practical deployment because of their efficiency, controllability, and low cost. This explicit layout analysis process is critical. In this work, we focus on the second setting.

\begin{figure}[t]
    \centering
    \includegraphics[width=\linewidth]{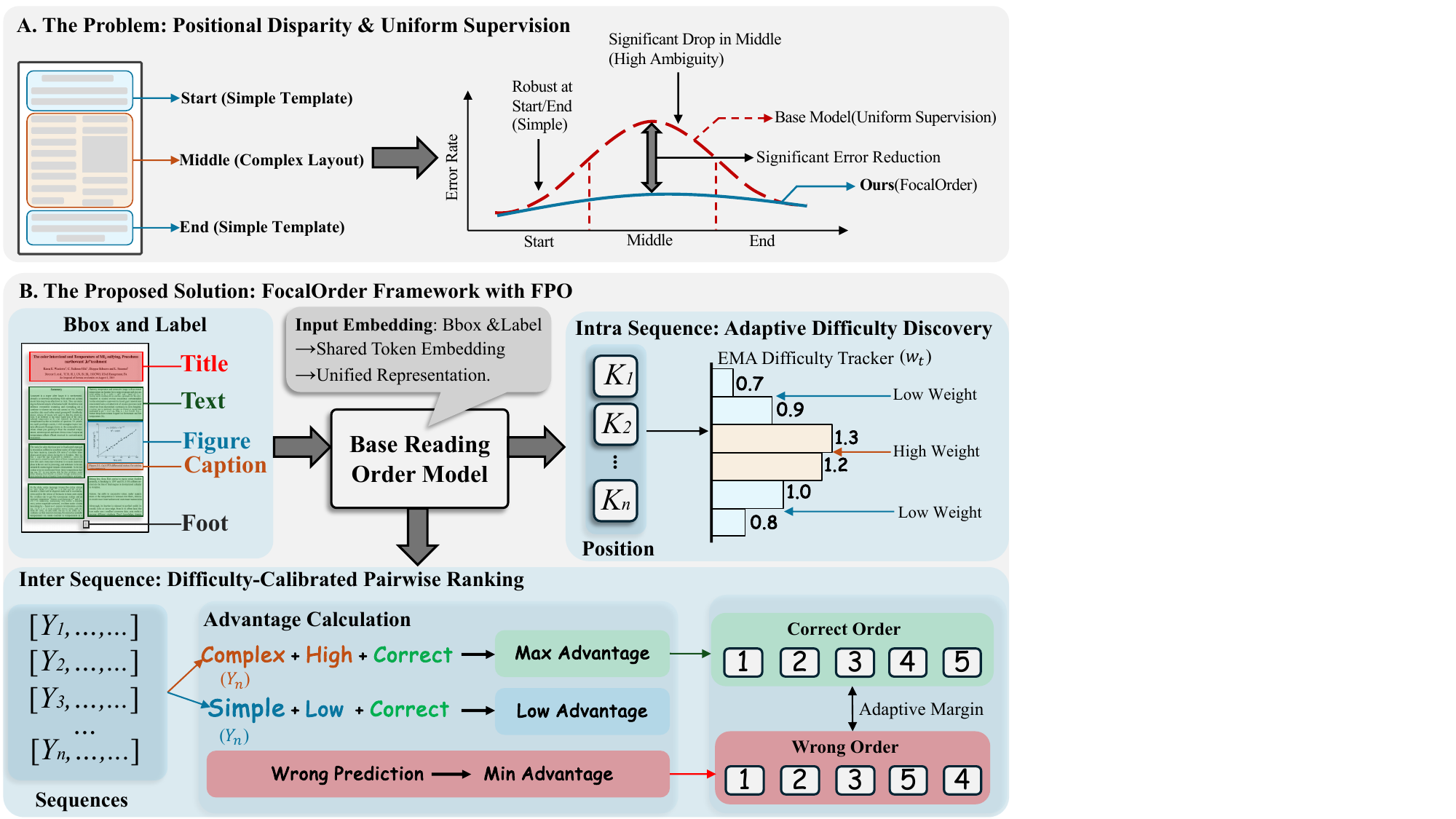}
    \setlength{\abovecaptionskip}{-8pt}
    \caption{Illustration of detector-to-parser handoff failures on a page. A dense page may produce multiple overlapping layout candidates around the same content region. Without suppression, duplicated candidates may be passed to the parser, causing content redundancy and an incorrect reading sequence. With heuristic NMS, an incomplete survivor may be retained while a better-localized alternative is removed, leading to content loss and an unstable parser input.}
    \Description{A four-panel diagram comparing ground truth, overlapping layout candidates, and parser outputs with and without non-maximum suppression. Overlapping candidates cause duplicated content and reading-order errors without suppression, while heuristic suppression may keep an incomplete region and remove a better-localized one, leading to content loss.}
    \label{fig:handoff_failure}
\end{figure}

A fundamental challenge in these pipeline systems arises from the discrepancy between the raw output of layout detectors and the increasingly structured input requirements of downstream parsers. While detectors typically generate an extensive and redundant pool of region hypotheses, downstream parsers require a refined, serialized subset of layout instances to operate effectively. The resulting collection of retained regions, organized into sequence, constitutes the parser interface. Traditionally, this interface is constructed using heuristic filters, such as Non-Maximum Suppression (NMS), alongside rule-based serialization. We define the critical process of transforming raw detections into this finalized interface as the detector-to-parser handoff. Crucially, this handoff is not merely a trivial data transfer, but rather a highly complex structural decision-making step that directly dictates the final parsing quality.

However, this handoff is often unstable on dense pages characterized by overlapping regions, duplicated candidates, ambiguous boundaries, and visually similar text blocks. Around a single true layout element, the detector may produce multiple competing hypotheses with slightly different bounding boxes or confidence patterns. Because rigid filters like NMS dictate which hypotheses remain while the reading order is often inferred separately, the retained layout regions can become misaligned with their assigned sequence. As illustrated in Figure \ref{fig:handoff_failure}, this instability leads to severe parsing errors, such as content redundancy or information loss.

This observation reveals an underexplored systems issue in pipeline document parsing. Existing work has improved layout detection, reading-order modeling, and multimodal parsing, yet less attention has been paid to whether the retained instance set and its order are jointly stabilized before handoff to the parser. Rather than treating detection, filtering, and ordering as loosely connected stages, we focus on the structure delivered to the parser.

Based on these observations, we introduce a lightweight parser-oriented structural refinement stage on top of a DETR-style \cite{detr} detector. Starting from the detector hypothesis pool, the proposed module performs set-level reasoning over detector query features, semantic cues, box geometry, and image evidence. From the same refined structural state, it jointly resolves box localization, instance retention, and parser input order before handoff. We further introduce retention-oriented supervision and a difficulty-aware ordering objective to better align the retained instance set and its serialized order with the final parser input.

Experiments on OmniDocBench, D4LA, DocLayNet, and Real5-OmniDocBench validate the effectiveness of our design. Our method improves page-level layout quality on challenging benchmarks and yields substantial gains in reading-order-sensitive parsing. When integrated with a fixed PaddleOCR-VL-1.5 backend, it achieves a Reading Order Edit of 0.024 on OmniDocBench, substantially reducing sequence mismatch at the detector-to-parser handoff.

Our contributions are summarized as follows:
\begin{itemize}
\item We propose a lightweight parser-oriented structural refinement component for pipeline document parsing, which transforms the detector hypothesis pool into a cleaner and better-ordered layout set before parser handoff.

\item Built on this component, we develop a joint refinement framework that unifies localization, retention, and parser input ordering, with retention-oriented supervision and a difficulty-aware ordering objective.

\item Extensive experiments on multiple public benchmarks show that our method consistently improves page-level layout quality and yields particularly strong gains in reading-order-sensitive parsing under a fixed downstream parser.
\end{itemize}

\section{Related Work}

\subsection{End-to-End OCR Parsers}

Contemporary document parsing systems can be broadly categorized into end-to-end optical character recognition (OCR) parsers and OCR systems equipped with an explicit Document Layout Analysis (DLA) front-end. In contrast to traditional multi-stage pipelines, earlier document understanding systems, such as TRIE \cite{trie}, explored coupling text reading and downstream extraction within a unified framework. Recent large-scale parsers further advance this paradigm by directly converting page images into structured outputs, bypassing a discrete DLA stage. Representative systems include FireRedOCR \cite{fireredocr}, HunyuanOCR \cite{hunyuanocr}, DeepSeekOCR \cite{deepseekocr}, olmOCR \cite{olmocr}, DREAM \cite{dream} and POINTS-Reader \cite{pointsreader}. Despite variations in scale and design, these models predominantly treat layout reasoning and reading-order recovery as implicit components of end-to-end document conversion. Because our work focuses on the structural handoff between a detector and a downstream parser, we discuss this trajectory only briefly.

\subsection{OCR Systems with Explicit DLA Front-ends}

A substantial body of OCR systems continues to rely on explicit DLA front-ends, wherein layout regions are initially detected and subsequently passed to downstream recognition, reading-order recovery, or document conversion modules. Contemporary layout detectors predominantly formulate layout analysis as a multi-object detection task. Within this pipeline architecture, several studies focus on enhancing region localization across diverse document types, as seen in DocLayout-YOLO \cite{doclayoutyolo}, PP-DocLayout \cite{ppdoclayout}, and PARL \cite{liu2026parl}. A complementary line of research addresses parser-side ordering and serialization. LayoutReader \cite{layoutreader} predicts reading orders derived from text and layout cues, while recent work such as FocalOrder \cite{focalorder} further explores reading-order modeling under structurally ambiguous layouts. Although flexible and efficient, these pipelines remain susceptible to missed regions, duplicate predictions, and cross-stage error propagation. Another research trajectory strengthens explicit DLA front-ends via enhanced document representations and structure-aware modeling. DiT \cite{dit} demonstrates that document-domain pretraining can substantially benefit downstream layout analysis, while LayoutLMv3 \cite{layoutlmv3} unifies text, image, and layout modeling within a generalized multimodal pretraining framework. Furthermore, mmLayout \cite{mmlayout} emphasizes multi-grained document structure modeling beyond local token-level features. At the system level, MonkeyOCR \cite{monkeyocr} and PP-StructureV3 \cite{ppstructurev3} integrate layout analysis with OCR and structured export within modular pipelines.

Subsequent studies model layout prediction and reading order more jointly. DLAFormer \cite{dlaformer} unifies multiple DLA subtasks within a Transformer-based framework. GraphLayoutLM \cite{graphlayoutlm} explores this trajectory by introducing graph-based layout modeling and sequence reordering for visually rich documents. PaddleOCR-VL adopts an integrated design via PP-DocLayoutV2 \cite{paddleocrvl}, combining layout detection with a lightweight pointer network for reading-order prediction. PP-DocLayoutV3 further advances this direction and is implemented in PaddleOCR-VL-1.5 \cite{paddleocrvl15} alongside related systems like GLM-OCR \cite{glmocr}. Beyond localization and ordering, GraphDoc \cite{graphdoc} and its DRGG baseline model richer document relations. Recent evaluations of Docling further suggest that parser-interface quality is not consistently reflected by COCO-style mAP metrics \cite{docling_layout_2025}. Compared to these methodologies, our focus is narrower and heavily system-oriented. Rather than redesigning the entire parsing stack, we investigate whether the retained instance set and its sequential order are jointly stabilized prior to parser handoff.

\begin{figure*}[t]
    \centering
    \includegraphics[width=\textwidth]{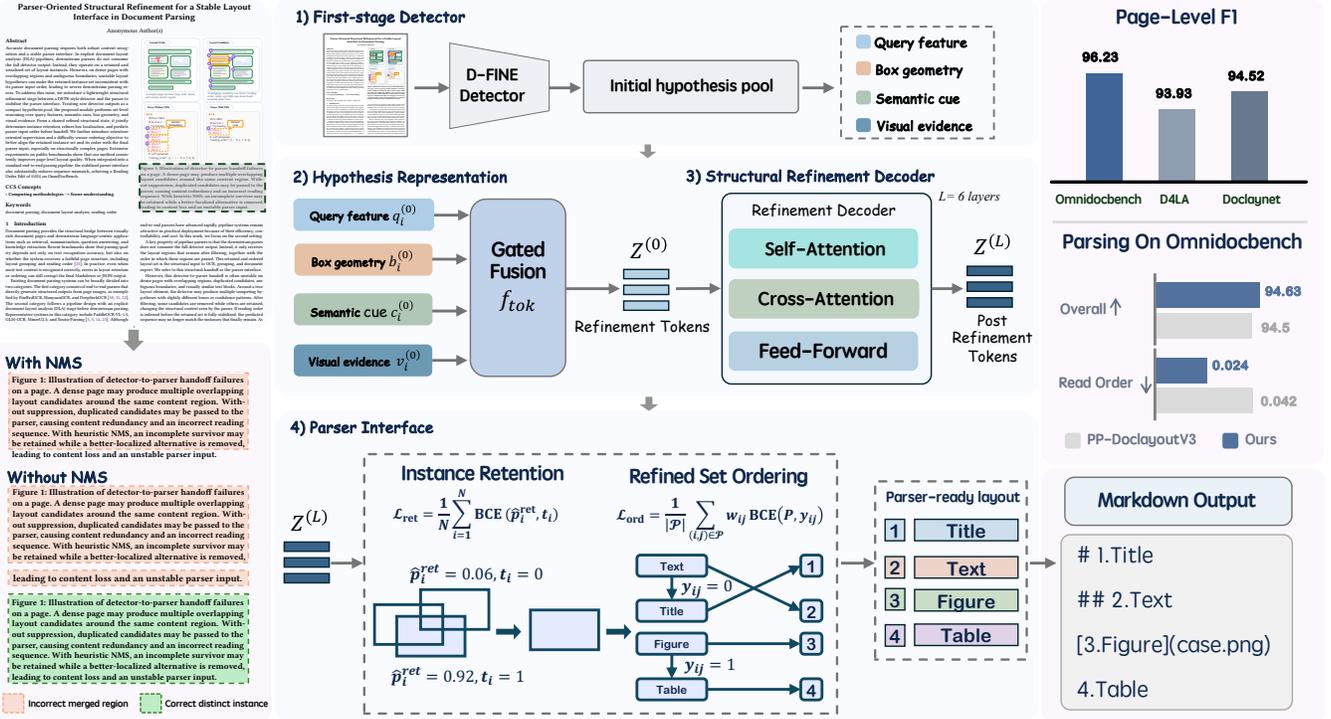}
    \setlength{\abovecaptionskip}{-8pt}
    \caption{Overview of the proposed parser-oriented structural refinement framework. Given a document image, a DETR-style detector first produces a compact pool of layout hypotheses. A lightweight refinement module then performs set-level reasoning over detector query features, semantic cues, box geometry, and image evidence, so that localization, instance retention, and parser input order are determined coherently from the same refined state. The downstream OCR-VL parser consumes a clean and well-ordered layout set, yielding more faithful Markdown/JSON output, particularly on structurally challenging pages.}
    \Description{An overview diagram of the proposed parser-oriented structural refinement framework. A detector first produces layout hypotheses from a document image. Query features, box geometry, semantic cues, and visual evidence are fused into refinement tokens, which are updated by a multi-layer refinement decoder. The refined states are then used for instance retention and ordering to form a parser-ready layout, improving page-level layout quality, reading order, and final Markdown output.}
    \label{fig:framework_overview}
\end{figure*}

A recent trend strengthens explicit DLA pipelines with multimodal or VLM-based modules. LayoutLMv3 \cite{layoutlmv3} provides an important foundation for joint modeling of text, layout, and image information in document understanding. Building on this direction, dots.ocr \cite{dotsocr} jointly learns layout detection, content recognition, and relational understanding within a single VLM. Dolphin-v2 \cite{dolphinv2} follows an analyze-then-parse design, in which the first stage predicts document elements together with reading order, and the second stage performs anchor-guided parsing. MinerU2.5 \cite{mineru25} separates global layout analysis from local high-resolution recognition in a coarse-to-fine framework, while Youtu-Parsing \cite{youtuparsing} uses a prompt-guided VLM for layout analysis and emphasizes reading-order restoration for complex documents. Later MonkeyOCR variants, such as MonkeyOCR v1.5 \cite{monkeyocrv15}, also move in this direction by replacing earlier lightweight modular combinations with stronger multimodal structure analysis and localized recognition. These systems improve structural modeling, but they also introduce higher inference and deployment cost. Our method is therefore positioned as a lightweight refinement layer for explicit DLA pipelines rather than as a replacement for heavyweight multimodal parsers.

\section{Method}
\label{sec:method}

We build a parser-oriented structural refinement module on top of D-FINE \cite{dfine}, a DETR-style detector, for pipeline document parsing. Rather than treating detector outputs as parser-ready instances, we regard them as a compact hypothesis pool that may contain duplicate, shifted, or structurally ambiguous candidates. The goal of our method is to transform this hypothesis pool into a stable parser interface before handoff. As shown in Figure~\ref{fig:framework_overview}, the method consists of four stages: first-stage detection, hypothesis representation, structural refinement decoding, and parser interface construction.

In a pipeline parser system, the downstream parser does not consume the full output of the layout detector. Instead, it consumes a parser interface, which consists of a serialized subset of retained layout instances. Let $\mathcal{Q}$ denote the detector hypothesis set, and let $\mathcal{S}\subseteq\mathcal{Q}$ denote the subset ultimately passed to the parser. The parser therefore depends on both the retained set $\mathcal{S}$ and its order. Conventional pipelines usually construct this interface through heuristic filtering and rule-based ordering, which can lead to unstable handoff decisions. Our method addresses this issue by refining the detector hypothesis pool and deriving retention and ordering from the same structural state.

\subsection{First-stage Detector}

Given a document page image $\mathbf{I}$, the first-stage detector produces a fixed set of $N$ layout hypotheses. We denote the detector output by:

{\small
\begin{equation}
\mathcal{Q}^{(0)}=
\left\{
\left(
\mathbf{q}_i^{(0)},\mathbf{b}_i^{(0)},\mathbf{c}_i^{(0)}
\right)
\right\}_{i=1}^{N},
\end{equation}
}where $\mathbf{q}_i^{(0)}$ is the detector query representation, $\mathbf{b}_i^{(0)}$ is the predicted box, and $\mathbf{c}_i^{(0)}$ is the semantic cue associated with the $i$-th hypothesis.

The backbone also provides multi-scale image features:

{\small
\begin{equation}
\mathcal{F}=\left\{\mathbf{F}^{(s)}\right\}_{s=1}^{S},
\end{equation}
}where $S$ is the number of feature levels.

\subsection{Hypothesis Representation}

Rather than passing detector hypotheses directly to the downstream parser, we convert each of them into a refinement token. Each token aggregates four sources of information: the detector query representation, box geometry, a compact semantic cue, and local visual evidence. This stage maps the detector hypothesis pool into a shared structural space in which competing candidates can be compared and refined before parser handoff.

Formally, we define the initial refinement token as:

{\small
\begin{equation}
\mathbf{z}_i^{(0)} = f_{\mathrm{tok}}
\left(
\mathbf{q}_i^{(0)},\mathbf{b}_i^{(0)},\mathbf{c}_i^{(0)},\mathbf{v}_i
\right),
\end{equation}
}where $\mathbf{v}_i$ denotes the box-conditioned visual feature extracted from $\mathcal{F}$, and $f_{\mathrm{tok}}(\cdot)$ is a learnable fusion module.

In implementation, box coordinates are represented in normalized $xyxy$ form and projected to the model dimension by an MLP. The semantic cue is implemented as a learnable class embedding, which provides a compact category prior rather than a full class-distribution signal. The visual cue is obtained by RoI-based multi-scale feature extraction over the predicted box region. We fuse these components with the detector query feature to form the initial refinement token. We use gated fusion so that the model can adaptively balance geometric, semantic, visual, and query-level evidence when constructing the shared representation.

\subsection{Structural Refinement Decoder}

Starting from the tokenized hypothesis pool $\mathcal{Z}^{(0)}=\{\mathbf{z}_i^{(0)}\}_{i=1}^{N}$, we apply a refinement decoder to perform set-level reasoning over the hypothesis set. This joint update is important because many handoff failures arise from structural relations among candidates, including duplication, overlap, competition, and ordering ambiguity.

We summarize the refinement process as follows:

{\small
\begin{equation}
\mathcal{Z}^{(L)} = f_{\mathrm{refine}}\!\left(\mathcal{Z}^{(0)},\mathcal{F}\right),
\end{equation}
}where $f_{\mathrm{refine}}(\cdot)$ denotes an $L$-layer refinement decoder.

In our implementation, the refinement decoder contains six layers. Each layer consists of self-attention, image-conditioned cross-attention, and a feed-forward network. The cross-attention module uses multi-scale deformable attention over four feature levels.

We further adopt iterative box refinement: each decoder layer predicts a box residual, and the refined box from the current layer is used as the reference box for the next layer. As a result, localization is improved progressively together with the latent hypothesis state. After the refinement layers, the model produces a shared refined state over the full hypothesis pool.

\subsection{Parser Interface}

The downstream parser does not consume the full refined hypothesis pool. Instead, it requires a retained set of layout instances together with their input order. We therefore construct the parser interface directly from the shared refined state.

From this state, the model predicts a category label, box localization, a retention score, and an ordering score for each hypothesis. The category label and box localization define the layout prediction of each candidate. The retention score determines whether the candidate remains in the parser input set, and the ordering score ranks the retained instances before they are passed downstream.

The retention prediction is implemented by a retention head on top of the refined hypothesis states. Instead of applying a fixed IoU-based suppression rule, this head learns which candidates should remain in the final parser input set, including difficult cases involving overlap and structural competition. This shared-state design is central to our method. Localization, retention, and ordering are all derived from the same refined hypothesis state, so that ordering is defined over the instance set that survives refinement.

\subsection{Training Objectives}

Training is performed on the refined hypothesis pool. We retain the detector's set-prediction objective for category prediction and box regression. On top of this objective, we introduce two additional supervision signals.

Detection loss.
Following D-FINE, we apply one-to-one Hungarian matching between refined predictions and the ground-truth set $\mathcal{Y}$. The detection loss is defined as:

{\small
\begin{equation}
\mathcal{L}_{\mathrm{det}} =
\lambda_{\mathrm{cls}}\mathcal{L}_{\mathrm{cls}}+
\lambda_{\mathrm{l1}}\mathcal{L}_{\mathrm{l1}}+
\lambda_{\mathrm{giou}}\mathcal{L}_{\mathrm{giou}},
\end{equation}
}where $\mathcal{L}_{\mathrm{cls}}$ is the classification loss, $\mathcal{L}_{\mathrm{l1}}$ is the L1 box regression loss, and $\mathcal{L}_{\mathrm{giou}}$ is the GIoU loss \cite{Giou2019}.

Retention-oriented supervision.
The first parser-oriented signal teaches the model which refined hypotheses should be retained in the final parser input set. Let $t_i\in\{0,1\}$ denote the retention target induced by the matching assignment, where $t_i=1$ if the $i$-th refined hypothesis is matched to a foreground ground-truth instance and $t_i=0$ otherwise. We define the retention loss as:

{\small
\begin{equation}
\mathcal{L}_{\mathrm{ret}}=
\frac{1}{N}\sum_{i=1}^{N}
\mathrm{BCE}\!\left(\hat{p}_i^{\mathrm{ret}},t_i\right),
\end{equation}
}where $\hat{p}_i^{\mathrm{ret}}$ is the predicted retention probability.

Ordering objective.
The second parser-oriented signal supervises parser input order over matched foreground instances. Let $\mathcal{I}$ denote the index set of refined hypotheses matched to foreground instances. For each pair $(i,j)$ with $i,j\in\mathcal{I}$ and $i\neq j$, let $y_{ij}=1$ if the instance matched to $i$ should appear earlier than the instance matched to $j$, and let $y_{ij}=0$ otherwise. We define the pairwise precedence probability as:

{\small
\begin{equation}
P(i\prec j)=\sigma\!\left(\hat{o}_j-\hat{o}_i\right),
\end{equation}
}where a smaller score indicates an earlier position.

The ordering loss is defined as:

{\small
\begin{equation}
\mathcal{L}_{\mathrm{ord}}=
\frac{1}{|\mathcal{P}|}
\sum_{(i,j)\in\mathcal{P}}
w_{ij}\,\mathrm{BCE}\!\left(P(i\prec j),y_{ij}\right),
\end{equation}
}where $\mathcal{P}$ is the set of valid foreground pairs and $w_{ij}$ is an optional pair weight.

Difficulty-aware weighting.
To emphasize difficult ordering transitions, we define the pair weight as:

{\small
\begin{equation}
w_{ij}=1+\gamma\log\!\left(1+n_{ij}^{\mathrm{mid}}\right),
\end{equation}
}where $n_{ij}^{\mathrm{mid}}$ is the number of other foreground ground-truth elements whose box centers fall inside the minimal axis-aligned rectangle spanned by the centers of the two matched instances, and $\gamma$ is a scalar coefficient.

Overall objective.
The full training objective is defined as:

{\small
\begin{equation}
\mathcal{L}=\mathcal{L}_{\mathrm{det}}+\lambda_{\mathrm{ret}}\mathcal{L}_{\mathrm{ret}}+\lambda_{\mathrm{ord}}\mathcal{L}_{\mathrm{ord}}.
\end{equation}
}

\subsection{Inference and Final Handoff}

At inference time, the model first refines the detector hypothesis pool and then constructs the final parser input in two steps. It first predicts retention scores for all refined hypotheses and keeps the candidates that should remain in the final parser input set. It then ranks only this retained subset according to the predicted ordering scores before passing the result downstream.

We define the retention score as:

{\small
\begin{equation}
s_i=\hat{p}_i^{\mathrm{ret}}\cdot \max_{c\in\{1,\dots,C\}}\hat{\pi}_{i,c},
\end{equation}
}where $\hat{\pi}_{i,c}$ is the refined foreground class probability for class $c$.

The retained subset is determined from this score without applying heuristic NMS. Instead, overlap handling is absorbed into the learned retention head, which resolves competition among overlapping candidates through set-conditioned structural reasoning. The ordering head predicts a scalar ordering score for each refined hypothesis, but only the retained subset participates in the final ranking step. The resulting retained and ordered layout set is then passed to the downstream OCR-VL parser.

\section{Experiments}
\label{sec:experiments}

\subsection{Experimental Setup}

Evaluation protocol and metrics.
We evaluate our method from two complementary perspectives. First, we assess page-level layout quality independently of any downstream parser on OmniDocBench \cite{omnidochbench}, D4LA \cite{vgt}, and DocLayNet \cite{doclaynet}, following the pageIoU-based protocol of MinerU2.5 \cite{mineru25} and reporting precision, recall, and F1. Here, D4LA is the benchmark introduced in Vision Grid Transformer. For D4LA and DocLayNet, we train only on their training splits and exclude the test partitions. Second, we evaluate whether a more stable detector-to-parser handoff improves end-to-end document parsing on OmniDocBench and Real5-OmniDocBench \cite{Real5OmniDocBench}, which are used exclusively for testing. To isolate the contribution of the DLA front-end, we fix the PaddleOCR-VL-1.5 backend and replace only its native layout analysis module with our method. On OmniDocBench, we report Text Block Edit and Reading Order Edit, where lower is better, together with Display Formula CDM, Table TEDS, Table TEDS Structure Only, and Overall, where higher is better. On Real5-OmniDocBench, we report Overall and Reading Order Edit on the full benchmark and its predefined subsets. Under this protocol, any performance differences directly reflect the quality of layout localization, retention, and reading-order recovery.

Implementation details.
Our framework is implemented in PyTorch. Unless otherwise specified, the first-stage detector and refinement module are trained jointly end to end. We use D-FINE-L as the detector and set the hypothesis pool size to $N=300$ in all experiments. We optimize the full model with AdamW on four NVIDIA RTX 4090 GPUs for 72 epochs, using a total batch size of 32, a base learning rate of $2.5 \times 10^{-4}$, and a weight decay of $1.25 \times 10^{-4}$ except for normalization parameters. The learning rate follows a MultiStepLR schedule. The refinement decoder has six layers, a hidden dimension of 256, and eight attention heads, with image-conditioned updates implemented by multi-scale deformable attention over four feature levels. For token construction, normalized box coordinates are projected by a two-layer MLP, semantic cues are represented by learnable class embeddings, and local visual evidence is extracted with multi-scale RoIAlign followed by pooling and linear projection. During training, the refinement module uses the same Hungarian assignment as the detector branch. During inference, we remove heuristic NMS, resolve candidate competition with the learned retention head, and rank only the retained subset by the predicted order scores before passing it to the downstream PaddleOCR-VL-1.5 parser. Additional implementation details and hyperparameter settings are provided in the Appendix.

\setlength{\textfloatsep}{16pt}
\begin{table}[t]
\centering
\setlength{\abovecaptionskip}{0.2cm}
\setlength{\belowcaptionskip}{-0.4cm}
\caption{PageIoU-based precision, recall, and F1 on OmniDocBench, D4LA, and DocLayNet. Best results are in \textbf{bold} and second-best results are \underline{underlined}.}
\label{tab:pageiou_results}
\footnotesize
\setlength{\tabcolsep}{2pt}
\begin{tabular}{lccccccccc}
\toprule
\multirow{2}{*}{Method} & \multicolumn{3}{c}{OmniDocBench} & \multicolumn{3}{c}{D4LA} & \multicolumn{3}{c}{DocLayNet} \\
\cmidrule(lr){2-4}\cmidrule(lr){5-7}\cmidrule(lr){8-10}
 & P & R & F1 & P & R & F1 & P & R & F1 \\
\midrule
DocLayout-YOLO & 92.30 & \textbf{97.70} & 94.10 & 82.60 & \underline{95.40} & 87.30 & 88.00 & 96.30 & 90.90 \\
PP-StructureV3 & 94.80 & 96.20 & 94.60 & 85.70 & 91.00 & 86.00 & 92.40 & 95.70 & 93.00 \\
MinerU2.5      & 95.80 & \underline{97.00} & 95.90 & 90.40 & 92.50 & 90.20 & \underline{92.80} & \underline{97.70} & \textbf{95.25} \\
PP-DocLayoutV3 & \underline{95.81} & 96.26 & \underline{96.03} & 89.04 & 90.40 & 89.71 & 91.62 & 95.46 & 93.50 \\
dots.ocr v1.5  & 94.97 & 96.21 & 95.59 & 91.03 & 94.57 & \underline{92.80} & 92.56 & 94.34 & 93.45 \\
Youtu-Parsing  & 92.99 & 96.07 & 94.53 & \underline{91.80} & 86.40 & 89.10 & \textbf{92.88} & 93.20 & 93.04 \\
Ours           & \textbf{95.97} & 96.50 & \textbf{96.23} & \textbf{92.04} & \textbf{95.90} & \textbf{93.93} & 91.49 & \textbf{97.76} & \underline{94.52} \\
\bottomrule
\end{tabular}
\end{table}

\begin{table*}[t]
\centering
\setlength{\abovecaptionskip}{0.2cm}
\setlength{\belowcaptionskip}{-0.4cm}
\caption{Results on Real5-OmniDocBench. Best results are in \textbf{bold} and second-best results are \underline{underlined}.}
\label{tab:real5_results}
\footnotesize
\setlength{\tabcolsep}{2pt}

\resizebox{\textwidth}{!}{%
\begin{tabular}{lc*{12}{c}}
\toprule
\multicolumn{1}{l}{\multirow{2}{*}{Method}} &
\multicolumn{1}{l}{\multirow{2}{*}{Params}} &
\multicolumn{2}{c}{Overall} &
\multicolumn{2}{c}{Scanning} &
\multicolumn{2}{c}{Warping} &
\multicolumn{2}{c}{Screen-Photography} &
\multicolumn{2}{c}{Illumination} &
\multicolumn{2}{c}{Skew} \\
\cmidrule(lr){3-4}\cmidrule(lr){5-6}\cmidrule(lr){7-8}\cmidrule(lr){9-10}\cmidrule(lr){11-12}\cmidrule(lr){13-14}
& & Score$\uparrow$ & Reading Order$\downarrow$ & Score$\uparrow$ & Reading Order$\downarrow$ & Score$\uparrow$ & Reading Order$\downarrow$ & Score$\uparrow$ & Reading Order$\downarrow$ & Score$\uparrow$ & Reading Order$\downarrow$ & Score$\uparrow$ & Reading Order$\downarrow$ \\
\midrule
\multicolumn{14}{c}{\textit{Pipeline Tools}} \\
\midrule
Marker-1.8.2     & -    & 60.10 & 0.367 & 70.27 & 0.238 & 58.98 & 0.390 & 63.65 & 0.325 & 66.31 & 0.337 & 41.27 & 0.543 \\
PP-StructureV3   & -    & 64.45 & 0.212 & 84.68 & 0.092 & 59.34 & 0.261 & 66.89 & 0.165 & 73.38 & 0.126 & 37.98 & 0.417 \\
\midrule
\multicolumn{14}{c}{\textit{Specialized VLMs}} \\
\midrule
DeepSeek-OCR 2   & 3B   & 73.01 & 0.153 & 89.59 & 0.056 & 66.53 & 0.209 & 71.65 & 0.157 & 76.02 & 0.122 & 61.28 & 0.221 \\
DeepSeek-OCR     & 3B   & 73.99 & 0.173 & 86.17 & 0.085 & 67.20 & 0.226 & 75.31 & 0.169 & 78.10 & 0.156 & 63.01 & 0.231 \\
MinerU2-VLM      & 0.9B & 76.95 & 0.140 & 83.60 & 0.091 & 73.73 & 0.173 & 78.77 & 0.123 & 80.51 & 0.123 & 68.16 & 0.191 \\
MonkeyOCR-pro-3B & 3.7B & 79.49 & 0.200 & 86.94 & 0.141 & 78.90 & 0.212 & 82.44 & 0.177 & 84.71 & 0.171 & 64.47 & 0.301 \\
Nanonets-OCR-s   & 3B   & 84.19 & 0.118 & 85.52 & 0.106 & 83.56 & 0.124 & 84.86 & 0.117 & 85.01 & 0.112 & 81.98 & 0.133 \\
dots.ocr         & 3B   & 86.38 & 0.085 & 86.87 & 0.081 & 86.01 & 0.093 & 87.18 & 0.079 & 87.57 & 0.076 & 84.27 & 0.094 \\
PaddleOCR-VL     & 0.9B & 85.54 & 0.099 & 92.11 & 0.048 & 85.97 & 0.092 & 82.54 & 0.107 & 89.61 & 0.055 & 77.47 & 0.193 \\
MinerU2.5        & 1.2B & 85.61 & 0.084 & 90.06 & 0.050 & 83.76 & 0.104 & 89.41 & \underline{0.053} & 89.57 & 0.062 & 75.24 & 0.151 \\
GLM-OCR          & 0.9B & 90.32 & 0.092 & 92.67 & 0.061 & \underline{90.68} & 0.100 & 91.75 & 0.070 & 91.12 & 0.071 & 85.39 & 0.156 \\
PaddleOCR-VL-1.5 & 0.9B & \textbf{92.05} & \underline{0.056} & \underline{93.43} & \underline{0.045} & \textbf{91.25} & \underline{0.063} & \underline{91.76} & 0.059 & \underline{92.16} & \underline{0.051} & \textbf{91.66} & \underline{0.061} \\
Ours             & 0.9B & \underline{91.63} & \textbf{0.036} & \textbf{94.33} & \textbf{0.027} & 89.34 & \textbf{0.040} & \textbf{93.46} & \textbf{0.036} & \textbf{94.07} & \textbf{0.030} & \underline{86.97} & \textbf{0.049} \\
\bottomrule
\end{tabular}%
}
\end{table*}

\subsection{Page-level Layout Evaluation}
\label{sec:pageiou_results}

We first evaluate page-level layout quality independently of downstream parsing. Table~\ref{tab:pageiou_results} shows that our method achieves the best F1 on OmniDocBench and D4LA, reaching 96.23 and 93.93, respectively. On OmniDocBench, it improves over PP-DocLayoutV3 by 0.20 F1 and over MinerU2.5 by 0.33 F1, showing that the proposed refinement remains effective against strong layout baselines. The gain is larger on D4LA, where our method surpasses dots.ocr v1.5 by 1.13 F1 and PP-DocLayoutV3 by 4.22 F1, suggesting that the benefit is more pronounced on structurally challenging pages. On DocLayNet, our method remains competitive with 94.52 F1. Although MinerU2.5 achieves a higher overall F1 on this benchmark, our model attains the highest recall of 97.76, indicating a stronger ability to preserve document elements. Overall, these results suggest that the proposed refinement improves the quality of the final retained layout set, especially on dense and complex pages.

\begin{table}[t]
\centering
\setlength{\abovecaptionskip}{0.2cm}
\setlength{\belowcaptionskip}{0.1cm}
\caption{Parsing results on OmniDocBench. Best results are in \textbf{bold} and second-best results are \underline{underlined}.}
\label{tab:omnidocbench_v15}
\tiny
\setlength{\tabcolsep}{1pt}
\begin{tabular}{lcccccc}
\toprule
Method & Overall$\uparrow$ & Text Edit$\downarrow$ & Formula CDM$\uparrow$ & Table TEDS$\uparrow$ & Table TEDS-S$\uparrow$ & Reading Order Edit$\downarrow$ \\
\midrule
\multicolumn{7}{c}{\textit{Pipeline Tools}} \\
\midrule
Marker-1.8.2     & 71.30 & 0.206 & 76.66 & 57.88 & 71.17 & 0.250 \\
PP-StructureV3   & 86.73 & 0.073 & 85.79 & 81.68 & 89.48 & 0.073 \\
\midrule
\multicolumn{7}{c}{\textit{Specialized VLMs}} \\
\midrule
Dolphin-v2       & 89.78 & 0.054 & 87.63 & 87.02 & 90.48 & 0.054 \\
OCRFlux-3B       & 74.82 & 0.193 & 68.03 & 75.75 & 80.23 & 0.202 \\
Mistral OCR      & 78.83 & 0.164 & 82.84 & 70.03 & 78.04 & 0.144 \\
POINTS-Reader    & 80.98 & 0.134 & 79.20 & 77.13 & 81.66 & 0.145 \\
olmOCR-7B        & 81.79 & 0.096 & 86.04 & 68.92 & 74.77 & 0.121 \\
Dolphin-1.5      & 83.21 & 0.092 & 80.78 & 78.06 & 84.10 & 0.080 \\
Nanonets-OCR-s   & 85.59 & 0.093 & 85.90 & 80.14 & 85.57 & 0.108 \\
MonkeyOCR-pro-3B & 88.85 & 0.075 & 87.25 & 86.78 & 90.63 & 0.128 \\
MinerU2.5        & 90.67 & 0.047 & 88.46 & 88.22 & 92.38 & 0.044 \\
DeepSeek-OCR 2   & 91.09 & 0.048 & 90.31 & 87.75 & 92.06 & 0.057 \\
FireRed-OCR-2B   & 92.94 & \underline{0.032} & 91.71 & 90.31 & 93.81 & 0.041 \\
MonkeyOCR v1.5   & 93.01 & 0.045 & 91.54 & 91.99 & 95.04 & 0.049 \\
Youtu-Parsing    & 93.22 & 0.045 & 93.19 & 91.15 & 95.43 & \underline{0.026} \\
dots.ocr v1.5    & 93.58 & \textbf{0.031} & 92.06 & 91.79 & 94.62 & 0.029 \\
Hunyuan-OCR      & 94.10 & 0.042 & \textbf{94.73} & 91.81 & - & - \\
PaddleOCR-VL-1.5 & 94.50 & 0.035 & \underline{94.21} & 92.76 & 95.79 & 0.042 \\
GLM-OCR          & \underline{94.62} & 0.040 & 93.90 & \textbf{93.96} & \textbf{96.39} & 0.044 \\
Ours             & \textbf{94.63} & 0.035 & 94.11 & \underline{93.28} & \underline{95.88} & \textbf{0.024} \\
\bottomrule
\end{tabular}
\end{table}

\subsection{End-to-end Parsing Results}
\label{sec:omnidocbench_v15_results}

We evaluate if the refined DLA front-end improves end-to-end parsing with a fixed downstream parser. Results on Real5-OmniDocBench and OmniDocBench are in Tables~\ref{tab:real5_results} and \ref{tab:omnidocbench_v15}, respectively.

Real5-OmniDocBench.
On Real5-OmniDocBench, our method achieves an overall score of 91.63 and the lowest Reading Order Edit of 0.036 in the evaluated comparison set. Although PaddleOCR-VL-1.5 attains a slightly higher overall score of 92.05, our method substantially reduces Reading Order Edit from 0.056 to 0.036. One plausible reason is that PaddleOCR-VL-1.5 uses PP-DocLayoutV3, whose region representation is better suited to heavily warped or skewed layouts than conventional two-point box formulations, which may partly explain its stronger overall scores on these subsets. Even so, our method achieves the best Reading Order Edit across all five subsets and attains the highest overall scores on Scanning, Screen-Photography, and Illumination. On Warping and Skew, although the overall score trails the strongest baseline, our method still preserves the best reading-order accuracy. These results suggest that the proposed refinement is especially effective for reading-order-sensitive parsing under realistic degradations, while remaining highly competitive in overall parsing quality.

OmniDocBench.
On OmniDocBench, our method reaches an overall score of 94.63, slightly above GLM-OCR (94.62) and 0.13 points higher than the PaddleOCR-VL-1.5 pipeline (94.50). The gain in Overall is modest, which is expected given the strength of the fixed backend and the fact that this aggregate score depends on multiple downstream factors beyond layout ordering alone. The clearest advantage appears in reading-order recovery: our method reduces Reading Order Edit to 0.024, outperforming all compared methods, including Youtu-Parsing (0.026), dots.ocr v1.5 (0.029), PaddleOCR-VL-1.5 (0.042), and GLM-OCR (0.044). This result is consistent with our main claim that the largest benefit comes from stabilizing the retained layout set before serialization. At the component level, our method also achieves competitive scores on Display Formula CDM (94.11), Table TEDS (93.28), and Table TEDS Structure Only (95.88), while maintaining Text Block Edit at 0.035.

\begin{table}[t]
\centering
\setlength{\abovecaptionskip}{0.2cm}
\setlength{\belowcaptionskip}{0.1cm}
\caption{Efficiency comparison of layout analysis under our implementation setting. Latency is measured in milliseconds per page. $N$ denotes the size of the input hypothesis pool.}
\label{tab:efficiency}
\footnotesize
\setlength{\tabcolsep}{6pt}
\begin{tabular}{lcc}
\toprule
Method & Latency (ms/page)$\downarrow$ & Params$\downarrow$ \\
\midrule
PP-DocLayoutV3                 & 154.45 & 32.48M \\
PP-DocLayoutV3 + NMS           & 154.52 & 32.48M \\
D-FINE                         & 119.55 & 30.68M \\
D-FINE + NMS                   & 119.70 & 30.68M \\
D-FINE + LayoutReader + NMS    & 199.63 & 400M \\
Ours ($N=300$)                 & 148.34 & 61.57M \\
Ours ($N=500$)                 & 151.54 & 61.57M \\
\midrule
MinerU2.5                      & 543.00 & 1.2B \\
Youtu-Parsing                  & 1982.88 & 3B \\
\bottomrule
\end{tabular}
\end{table}

\subsection{Efficiency Analysis}

We evaluate efficiency in terms of inference latency and parameter count. As shown in Table~\ref{tab:efficiency}, our method introduces moderate overhead relative to detector-only baselines while remaining substantially lighter than heavyweight multimodal systems. Compared with PP-DocLayoutV3 + NMS, our latency remains competitive while delivering stronger layout quality and reading-order performance. Compared with D-FINE + NMS, latency increases from 119.70 to 148.34 ms/page, but page-level quality and reading-order performance improve consistently. Increasing the hypothesis pool from $N=300$ to $N=500$ raises latency only slightly, from 148.34 to 151.54 ms/page, suggesting good scalability. Compared with the decoupled baseline D-FINE + LayoutReader + NMS, our method is both smaller and faster, requiring 61.57M parameters and 148.34 ms/page versus 400M and 199.63 ms/page. It also remains much lighter than systems such as MinerU2.5 and Youtu-Parsing.

\subsection{Ablation Studies}
\label{sec:ablation_studies}

We study three questions: 1) whether learned refinement is more effective than heuristic filtering; 2) which components of the proposed framework contribute most to performance; and 3) how broadly the refined layout interface can benefit downstream parsing.

NMS-free structural refinement.
We first compare learned structural refinement with heuristic post-processing under the pageIoU protocol. Table~\ref{tab:ablation_nms} reports results for two DETR-style front-ends, PP-DocLayoutV3 and D-FINE, with and without NMS. Although these detectors use one-to-one set prediction, their outputs can still contain redundant or highly overlapping hypotheses on dense and ambiguous pages. Under pageIoU evaluation, such duplicates mainly hurt precision. Removing NMS generally increases recall but reduces precision, while heuristic NMS partially restores the precision-recall balance. However, NMS only changes which boxes remain; it does not jointly resolve localization, retention, and ordering. Built on D-FINE, our method achieves the best F1 on all three datasets, with gains of +0.14 on D4LA, +2.51 on DocLayNet, and +0.49 on OmniDocBench over D-FINE + NMS. These results show that learned refinement is more effective than heuristic suppression for producing the final layout set.

\begin{table}[t]
\centering
\setlength{\abovecaptionskip}{0.2cm}
\setlength{\belowcaptionskip}{-0.1cm}
\caption{Ablation on NMS-free structural refinement. Best results are in \textbf{bold} and second-best results are \underline{underlined}.}
\label{tab:ablation_nms}
\footnotesize
\setlength{\tabcolsep}{1.5pt}
\begin{tabular}{lccccccccc}
\toprule
\multirow{2}{*}{Method} & \multicolumn{3}{c}{D4LA} & \multicolumn{3}{c}{DocLayNet} & \multicolumn{3}{c}{OmniDocBench} \\
\cmidrule(lr){2-4}\cmidrule(lr){5-7}\cmidrule(lr){8-10}
 & P & R & F1 & P & R & F1 & P & R & F1 \\
\midrule
PP-DocLayoutV3         & 82.91 & 93.96 & 88.09 & 87.01 & 95.50 & 91.06 & 90.25 & 96.68 & 93.35 \\
PP-DocLayoutV3 + NMS   & 89.04 & 90.40 & 89.71 & \underline{91.62} & 95.46 & \underline{93.50} & \underline{95.81} & 96.26 & \underline{96.03} \\
D-FINE                 & 87.41 & \textbf{97.62} & 92.23 & 80.91 & \underline{97.92} & 88.61 & 93.70 & \textbf{97.04} & 95.34 \\
D-FINE + NMS           & \underline{91.76} & 95.92 & \underline{93.79} & 86.97 & 97.68 & 92.01 & 95.52 & 95.97 & 95.74 \\
Ours                   & \textbf{92.04} & 95.90 & \textbf{93.93} & 91.49 & \textbf{97.76} & \textbf{94.52} & \textbf{95.97} & 96.50 & \textbf{96.23} \\
\bottomrule
\end{tabular}
\end{table}

\begin{table}[t]
\centering
\setlength{\abovecaptionskip}{0.15cm}

\caption{Class-wise F1 on OmniDocBench. Best results are in \textbf{bold} and second-best results are \underline{underlined}.}
\label{tab:classwise_omnidoc}
\scriptsize
\setlength{\tabcolsep}{6pt}
\begin{tabular}{lccccc}
\toprule
Method & Textual & Image & Table & Equation & Page\_margins \\
\midrule
D-FINE               & 0.967 & 0.947 & 0.968 & 0.885 & 0.857 \\
D-FINE + NMS         & \underline{0.971} & \underline{0.956} & \underline{0.972} & 0.901 & 0.873 \\
PP-DocLayoutV3 + NMS & 0.966 & \textbf{0.961} & 0.966 & \underline{0.904} & \underline{0.902} \\
Ours                 & \textbf{0.972} & 0.956 & \textbf{0.975} & \textbf{0.932} & \textbf{0.933} \\
\bottomrule
\end{tabular}
\end{table}

Class-wise analysis on OmniDocBench.
To diagnose where the gain comes from, we report class-wise F1 on OmniDocBench in Table~\ref{tab:classwise_omnidoc}. Heuristic NMS improves D-FINE across most categories, showing that duplicate suppression is useful even for DETR-style detectors on dense pages. The proposed refinement yields further gains in Textual, Table, Equation, and Page\_margins, while PP-DocLayoutV3 + NMS remains stronger on Image. In particular, compared with PP-DocLayoutV3 + NMS, our method shows clearer advantages on Equation and Page\_margins. These categories are more sensitive to fragmented survivors and incomplete region retention under heuristic filtering. By contrast, natural images are often more compact and visually separable, which may reduce the relative benefit of our refinement. Overall, this pattern suggests that the main advantage of the proposed design lies in structurally sensitive categories, where parser-facing layout quality depends more critically on stable retention than on local suppression alone.

Core comparison and component analysis.
We next perform a compact comparison on OmniDocBench to clarify the roles of parser-oriented refinement, retention-oriented supervision, and difficulty-aware ordering. Since the Overall score on OmniDocBench does not include reading order, variants that mainly affect ordering should be compared primarily through Reading Order Edit.

In Table~\ref{tab:core_ablation}, heuristic overlap suppression is helpful but limited: D-FINE + NMS improves F1 and Overall over raw D-FINE, yet it does not address the mismatch between instance filtering and parser input ordering. To examine whether a decoupled reading-order model can solve this problem, we attach LayoutReader after detection. We choose LayoutReader as a representative, publicly available reading-order model in explicit DLA pipelines. However, D-FINE + LayoutReader + NMS produces the same F1 and Overall as D-FINE + NMS, while yielding a much worse Reading Order Edit of 0.175. This suggests that ordering predicted after heuristic filtering remains sensitive to errors in the retained instance set. In contrast, our unified refinement achieves the best overall result. The ablations show that retention-oriented supervision is important for instance selection. Removing $\mathcal{L}_{\mathrm{ret}}$ reduces F1 from 96.23 to 95.41, lowers Overall from 94.63 to 90.84, and increases Reading Order Edit from 0.024 to 0.084. Difficulty-aware weighting mainly affects ordering consistency: removing it leaves F1 and Overall unchanged, but worsens Reading Order Edit from 0.024 to 0.061.

\begin{table}[t]
\centering
\setlength{\abovecaptionskip}{0.2cm}
\setlength{\belowcaptionskip}{-0.2cm}
\caption{Compact core comparison and objective ablation on OmniDocBench. N/A indicates no reading-order prediction.}
\label{tab:core_ablation}
\tiny
\setlength{\tabcolsep}{8pt}
\renewcommand{\arraystretch}{1.08}
\begin{tabular}{lccc}
\toprule
Variant & OmniDocBench F1$\uparrow$ & Overall$\uparrow$ & Reading Order Edit$\downarrow$ \\
\midrule
D-FINE                            & 95.34 & 89.97 & N/A \\
D-FINE + NMS                      & 95.74 & 93.24 & N/A \\
D-FINE + Soft-NMS                 & 95.79 & 93.29 & N/A \\
D-FINE + LayoutReader + NMS       & 95.74 & 93.24 & 0.175 \\
D-FINE + LayoutReader + Soft-NMS  & 95.79 & 93.29 & 0.171 \\
Ours w/o $\mathcal{L}_{\mathrm{ret}}$ & 95.41 & 90.84 & 0.084 \\
Ours w/o difficulty-aware         & 96.23 & 94.63 & 0.061 \\
Ours                              & \textbf{96.23} & \textbf{94.63} & \textbf{0.024} \\
\bottomrule
\end{tabular}
\vspace{-0.35cm}
\end{table}

\begin{figure*}[t]
    \centering
    \includegraphics[width=\textwidth]{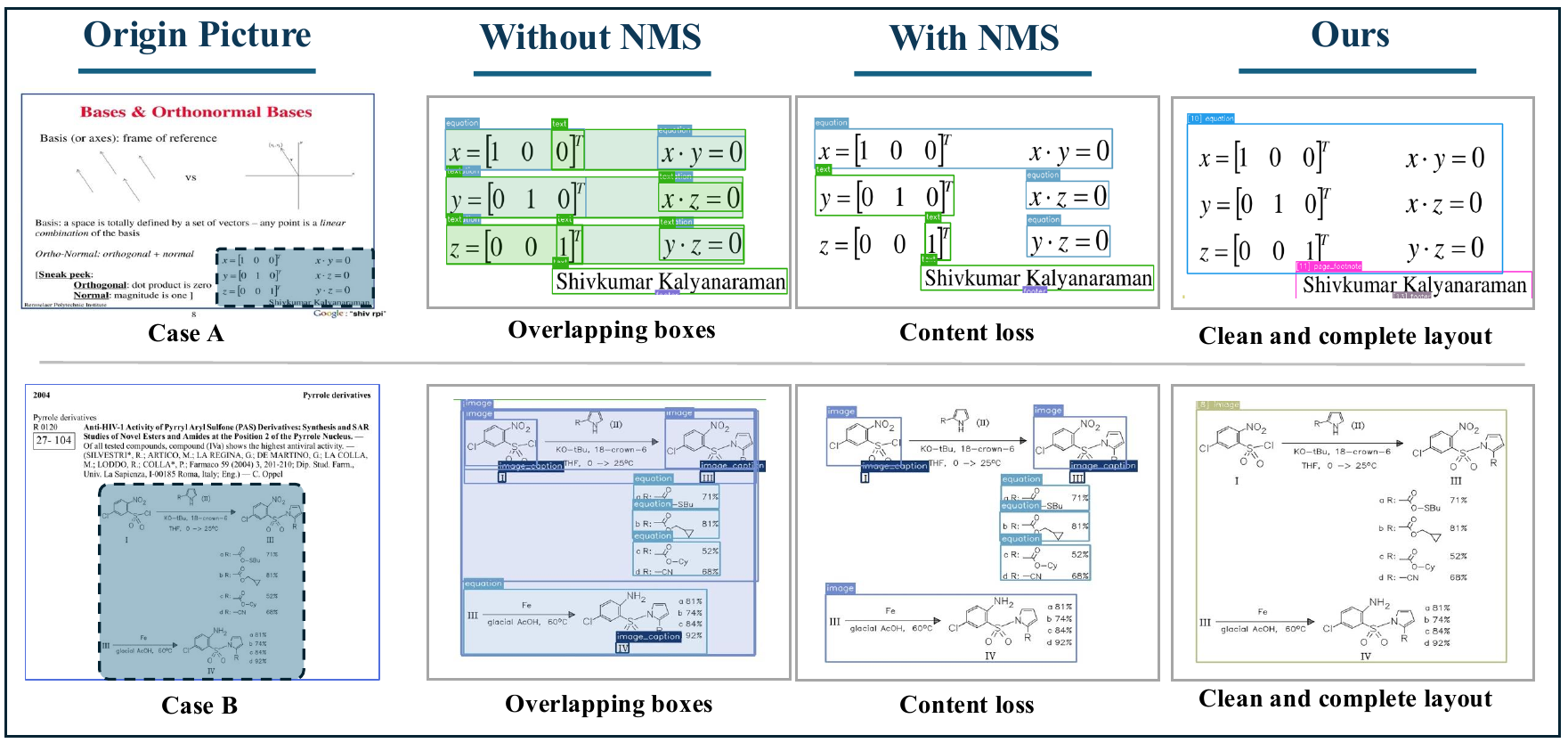}
    \setlength{\abovecaptionskip}{-6pt}
    \caption{Qualitative comparison on two challenging layouts (Case A: equations, Case B: chemical schemes). Our method avoids the severe redundancy of raw outputs and the content loss caused by traditional NMS, yielding clean and complete structures.}
    \Description{A two-row qualitative comparison of layout predictions on an equation page and a chemical-scheme page. Without NMS, multiple overlapping boxes are produced. With heuristic NMS, some valid content is removed and the retained layout becomes incomplete. The proposed method produces clean, complete, and well-aligned layout regions in both cases.}
    \label{fig:qualitative_cases}
\end{figure*}

\subsection{Qualitative Analysis}

Figure~\ref{fig:qualitative_cases} presents two cases that illustrate the difference between heuristic NMS and the proposed refinement module on complex document layouts. In both Case A, which contains mathematical equations, and Case B, which contains chemical schemes, the raw detector outputs without NMS include multiple overlapping candidates around the content regions, which would introduce redundant inputs to the downstream parser. Heuristic NMS reduces this duplication, but it may suppress structurally useful regions. In Case A, NMS fragments the equation region and removes part of the valid content, leaving only a partial instance. In Case B, NMS breaks the chemical figure into several small fragments instead of preserving it as a coherent layout element. By contrast, our refinement module produces a more complete and coherent layout set. It merges the multi-line equation block in Case A and preserves the structure of the chemical scheme in Case B with a better aligned bounding box.

\begin{table}[t]
\centering
\setlength{\abovecaptionskip}{0.2cm}
\caption{Exploratory compatibility study on OmniDocBench. The first row reports the original result, and the second row reports the result after incorporating our module.}
\label{tab:ablation_integration}
\footnotesize
\setlength{\tabcolsep}{5pt}
\renewcommand{\arraystretch}{1.1}
\begin{tabular}{llcc}
\toprule
Pipeline & Variant & Reading Order Edit$\downarrow$ & Overall$\uparrow$ \\
\midrule
\multirow[c]{2}{*}{MinerU2.5}
& Original & 0.044 & 90.67 \\
\cmidrule(lr){2-4}
& + Ours & 0.032 ($\downarrow$27.3\%) & 94.03 ($\uparrow$3.7\%) \\
\midrule
\multirow[c]{2}{*}{DeepSeek-OCR 2}
& Original & 0.057 & 91.09 \\
\cmidrule(lr){2-4}
& + Ours & 0.034 ($\downarrow$40.4\%) & 92.48 ($\uparrow$1.5\%) \\
\midrule
\multirow[c]{2}{*}{FireRed-OCR-2B}
& Original & 0.041 & 92.94 \\
\cmidrule(lr){2-4}
& + Ours & 0.029 ($\downarrow$29.3\%) & 93.98 ($\uparrow$1.1\%) \\
\midrule
\multirow[c]{2}{*}{GLM-OCR}
& Original & 0.044 & 94.62 \\
\cmidrule(lr){2-4}
& + Ours & 0.026 ($\downarrow$40.9\%) & 94.49 ($\downarrow$0.1\%) \\
\midrule
\multirow[c]{2}{*}{PaddleOCR-VL-1.5}
& Original & 0.042 & 94.50 \\
\cmidrule(lr){2-4}
& + Ours & 0.024 ($\downarrow$42.9\%) & 94.63 ($\uparrow$0.1\%) \\
\bottomrule
\end{tabular}
\end{table}

\subsection{Compatibility with Downstream OCR}

We further examine whether the refined layout interface can benefit different downstream backends. Table~\ref{tab:ablation_integration} compares the original system performance with the performance after incorporating our module. The evaluated baselines employ diverse layout strategies: some rely on lightweight DLA models (e.g., PaddleOCR-VL-1.5 and GLM-OCR), some use VLMs for layout reasoning (e.g., MinerU2.5), and others operate end-to-end without an explicit DLA stage. To unify the integration, we replace their native layout analysis, or introduce one for the end-to-end models, with our refined DLA front-end. Specifically, we use our module to extract and order the retained layout regions, then crop the corresponding image patches and feed them sequentially into the respective backends for recognition. Across all five systems, our module consistently improves reading-order quality, yielding absolute reductions from 0.012 to 0.023 in Reading Order Edit. The Overall score also improves in four of the five cases. GLM-OCR is the only exception, showing a small Overall drop from 94.62 to 94.49, although its Reading Order Edit still improves substantially from 0.044 to 0.026. This gap may reflect compatibility differences between our DLA front-end and the input assumptions of heterogeneous downstream backends. Even in this case, the system remains highly competitive overall while achieving much stronger reading-order accuracy.

\section{Conclusion}

This paper presents a parser-oriented structural refinement framework for explicit Document Layout Analysis. The proposed method improves the detector-to-parser handoff by refining first-stage detections before downstream parsing, so that localization, retention, and reading order are resolved more coherently. Experiments on OmniDocBench, D4LA, DocLayNet, and Real5-OmniDocBench validate the effectiveness of this design. Our method achieves strong page-level layout quality on OmniDocBench and D4LA and delivers competitive end-to-end parsing results with a fixed PaddleOCR-VL-1.5 backend, with the clearest gains appearing in reading-order recovery. These findings suggest that, in explicit DLA pipelines, improving the parser interface is a practical and effective way to strengthen document parsing. The proposed refinement is lightweight and modular, further highlighting the importance of parser-facing interface quality. At the same time, the current reading-order formulation and evaluation protocols mainly assume a fixed linear reading order. In practice, however, some documents admit multiple valid reading sequences, especially in structurally complex cases. This limitation suggests that current benchmarks do not fully reflect practical document parsing quality in such scenarios. Future work investigates more reliable modeling and evaluation protocols for reading order and structural ambiguity in complex documents.

\clearpage
\bibliographystyle{ACM-Reference-Format}
\bibliography{sample-base}

\end{document}